\documentclass{article}
\pdfoutput=1

\PassOptionsToPackage{numbers, compress}{natbib}

\usepackage[final]{neurips_2022}
\usepackage{graphicx}

\usepackage[utf8]{inputenc} 
\usepackage[T1]{fontenc}    
\usepackage{hyperref}       
\usepackage{url}            
\usepackage{booktabs}       
\usepackage{amsfonts}       
\usepackage{nicefrac}       
\usepackage{microtype}      
\usepackage{xcolor}         

\title{A New Graph Node Classification Benchmark: Learning Structure from Histology Cell Graphs}

\author{%
  Claudia Vanea\textsuperscript{1, 2}, 
  Jonathan Campbell\textsuperscript{1, 2, 3}, 
  Omri Dodi\textsuperscript{4}, 
  Liis Salumäe\textsuperscript{5}, 
  Karen Meir\textsuperscript{4},\\
  \textbf{Drorith Hochner-Celnikier\textsuperscript{4}, 
  Hagit Hochner\textsuperscript{6}, 
  Triin Laisk\textsuperscript{7}, 
  Linda M. Ernst,\textsuperscript{8}}\\
  \textbf{Cecilia M. Lindgren}\thanks{Equal contribution.}  \textsuperscript{2}, 
  \textbf{Christoffer Nellåker}\footnotemark[1]  \textsuperscript{1, 2}\\
  \texttt{claudia.vanea@wrh.ox.ac.uk}, 
  \texttt{jcampbell@robots.ox.ac.uk},\\
  \texttt{omri.dodi@mail.huji.ac.il},
  \texttt{liis.salumae@kliinikum.ee},\\ 
  \texttt{karenm@hadassah.org.il}, 
  \texttt{hochner@hadassah.org.il},\\
  \texttt{hagit.hochner@mail.huji.ac.il}, 
  \texttt{triin.laisk@ut.ee},\\
  \texttt{lernst@northshore.org},
  \texttt{cecilia.lindgren@bdi.ox.ac.uk},\\
  \texttt{christoffer.nellaker@bdi.ox.ac.uk}\\
  Nuffield Department of Women's \& Reproductive Health, University of Oxford\textsuperscript{1}\\
  Big Data Institute, University of Oxford\textsuperscript{2}\\
  Visual Geometry Group, University of Oxford\textsuperscript{3}\\
  Faculty of Medicine, Hadassah Hebrew University Medical Center\textsuperscript{4}\\
  Department of Pathology, Tartu University Hospital\textsuperscript{5}\\
  Braun School of Public Health, Hebrew University of Jerusalem\textsuperscript{6}\\
  Institute of Genomics, University of Tartu\textsuperscript{7}\\
  Northshore University HealthSystem, University of Chicago\textsuperscript{8}
}

\begin{document}

\maketitle

\begin{abstract}
We introduce a new benchmark dataset, \emph{Placenta}, for node classification in an underexplored domain: predicting microanatomical tissue structures from cell graphs in placenta histology whole slide images. This problem is uniquely challenging for graph learning for a few reasons. Cell graphs are large (>1 million nodes per image), node features are varied (64-dimensions of 11 types of cells), class labels are imbalanced (9 classes ranging from 0.21\% of the data to 40.0\%), and cellular communities cluster into heterogeneously distributed tissues of widely varying sizes (from 11 nodes to 44,671 nodes for a single structure). Here, we release a dataset consisting of two cell graphs from two placenta histology images totalling 2,395,747 nodes, 799,745 of which have ground truth labels. We present inductive benchmark results for 7 scalable models and show how the unique qualities of cell graphs can help drive the development of novel graph neural network architectures.
\end{abstract}

\section{Introduction}

The initial graph representation learning benchmarks facilitated the creation of the first graph neural network (GNN) architectures. \emph{Cora}, \emph{Citeseer}, \emph{Pubmed} and \emph{NELL} \cite{yang_revisiting_2016}, \cite{carlson_toward_2010} drove the development of Graph Convolutional Network (GCN) \cite{kipf_semi-supervised_2017}, \emph{Reddit} was introduced alongside GraphSAGE \cite{hamilton_inductive_2017}, and Graph Attention Network (GAT) \cite{velickovic_graph_2018} achieved state-of-the-art on the protein-protein interaction (\emph{PPI}) dataset \cite{zitnik_predicting_2017}. However, it soon became apparent that these datasets were not sufficiently complex to challenge GNNs \cite{hu_open_2021, wu_simplifying_2019}. The datasets had small graphs or a small number of graphs (the largest dataset, \emph{Reddit}, had 232,965 nodes), they had potentially misleading train, validation and test splits \cite{shchur_pitfalls_2019}, and they were so well suited to graph learning that randomly initialised, untrained models could perform nearly as well as trained models \cite{velickovic_deep_2018}. In recent years, there has been an influx of graph benchmark datasets which address these limitations. They are larger and cover more domains, have realistic train, validation and test splits, and are challenging enough to have not been ‘solved’ by current GNN architectures \cite{hu_open_2021,chiang_cluster-gcn_2019,szklarczyk_string_2019}. 

Cell graphs in histology whole slide images (WSIs) are a domain which has been largely unexplored for graph learning and present a challenge to the graph learning community \cite{lu_capturing_2020,zhou_cgc-net_2019,gadiya_histographs_2019,chen_pathomic_2020,lu_slidegraph_2022}. Histology is the study of tissues and cells under a microscope. A histology WSI is a digitised, high-resolution scan of thin tissue samples showing individual cells, which can be upwards of 70GBs per image. All organs are composed of cells that collectively build up the structures and tissues comprising the body. Consequently, representing organs as cell graphs is a logical construction closely matching the true physical system. The value of cell graphs has been highlighted \cite{yener_cell-graphs_2017,javed_cellular_2020,sirinukunwattana_novel_2018}, but with today's computational power and automated aggregation methods, it can reach its full potential.

Here, we introduce \emph{Placenta}, a cell graph from two WSIs as a new benchmark dataset to facilitate GNN creation. More specifically, we present a node classification task of predicting one of 9 placenta microanatomical tissue structures to which cells belong (Appendix \ref{cell-tissue-types}). This problem is challenging from a graph learning perspective for a number of reasons. Firstly, each WSI contains a cell graph of >1 million nodes, the node features are defined from an imbalanced distribution of cell classes, and the tissue labels also have a high, but expected, class imbalance, from 0.21\% of the data to 40.0\% (Appendix \ref{distributions}). Secondly, unlike most graph datasets, node clusters are locally homophilic but spatially distributed throughout the graph, requiring models to understand same-type node clusters at different locations. Thirdly, as a WSI is only ever a small sample of the larger structure with arbitrary cut-offs in edges, out-of-plane cells, and optical and mechanical artefacts, the property of data incompleteness is inherently a feature of this data domain. Finally, placental tissue structures grow from one another in a tree-like manner \cite{benirschke_architecture_2012,ernst_placenta_2019}, so singular structures can contain multiple tissue classifications and these structures will vary in size (30-5,000 $\mu$m\textsuperscript{2}). Models must discriminate between small changes across small regions within node clusters but have a large enough receptive field to identify the largest of tissues. 

Relevant to the real world domain, developing automated methods which can quantify the structural tissue units within histology images would be an invaluable tool for the scientific community and, in future, may have applications to clinical histopathologic diagnoses and patient care. Furthermore, this dataset provides a benchmark for developing novel, scalable GNNs that can adapt to node feature and class label imbalance, that can understand both large and small node communities distributed throughout the graph, and that are robust to data incompleteness. These models could be applied to new domains with similar qualities, such in modelling disease epidemiology. 

\begin{table}
  \caption{Dataset characteristics for the intersection graph.}
  \label{dataset-characteristics}
  \centering
  \begin{tabular}{llllll}
    \toprule
    & \textbf{Nodes} & \textbf{Edges} & \textbf{Labels} & \textbf{Homophily} & \textbf{Degree} \\
    \midrule
    \textbf{Cell Graph 1} & 1,159,186 & 2,603,797 & 703,165 & 0.9868 & 5.4925 \\
    \textbf{Cell Graph 2} & 1,236,561 & 2,882,292 & 96,580  & 0.9984 & 5.6618 \\
    \bottomrule
  \end{tabular}
\end{table}

\section{Dataset Description}
\label{dataset-description}

\paragraph{Overview} At a high level, the learning task is to understand placental tissue structures as the composition of their cellular communities. Tissue structures in the placenta are precisely defined and distinguished from one another by their cellular composition \cite{benirschke_architecture_2012}. In a WSI these manifest as spatially clustered groups of cells, with corresponding tissue structures comprising 10s-10,000s of cell nodes. Inductive message-passing GNNs, which can aggregate node neighbourhood information irrespective of graph shapes and sizes are, in theory, a natural fit for this task.

\begin{figure}
  \centering
  \includegraphics[scale=0.78]{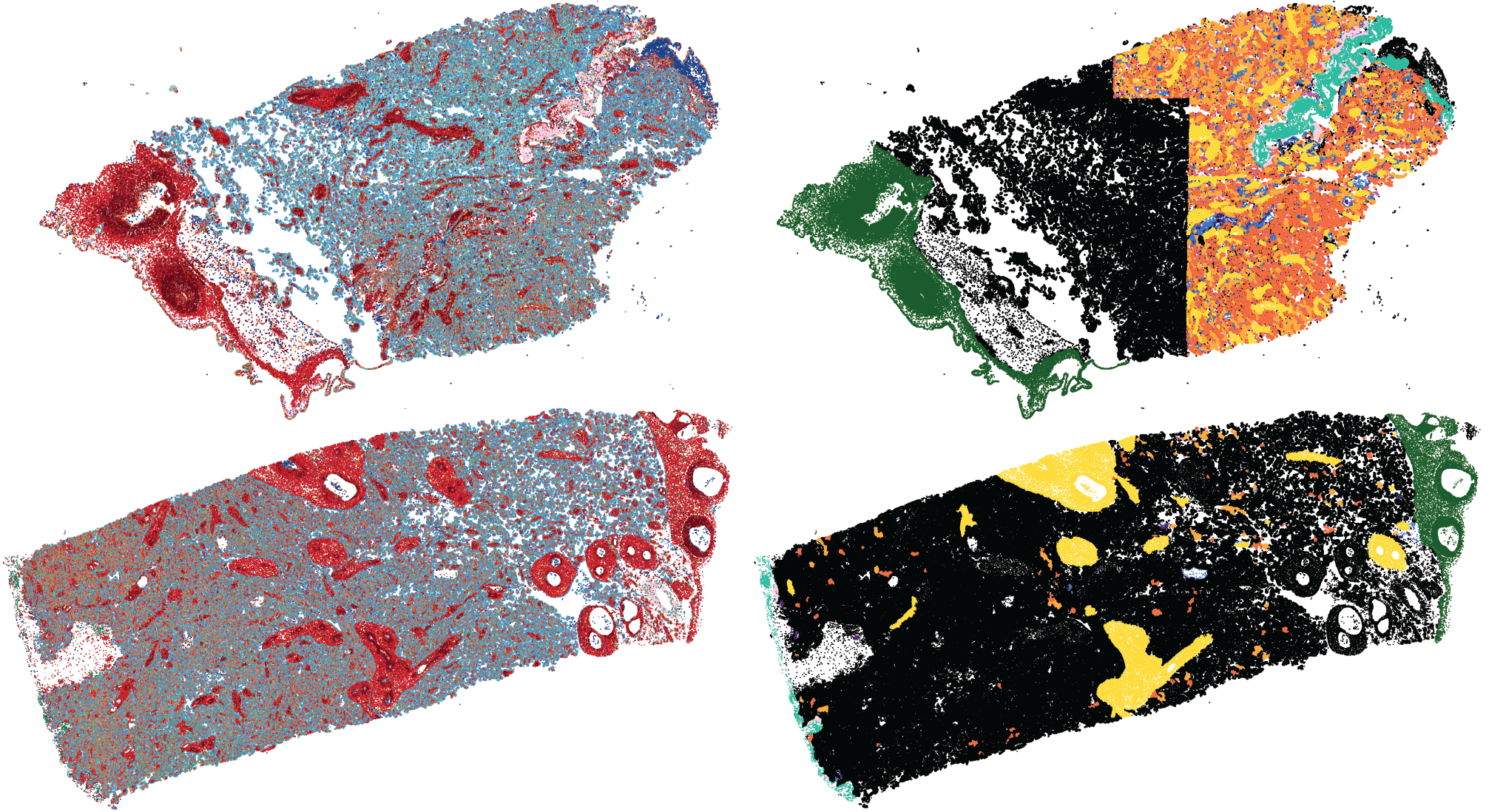}
  \caption{Cell graph nodes across from whole slide images (left) with corresponding labelled ground truth tissue nodes (right). Black tissue nodes are unlabelled.}
  \label{fig-1}
\end{figure}

\begin{figure}
  \centering
  \includegraphics[scale=0.80]{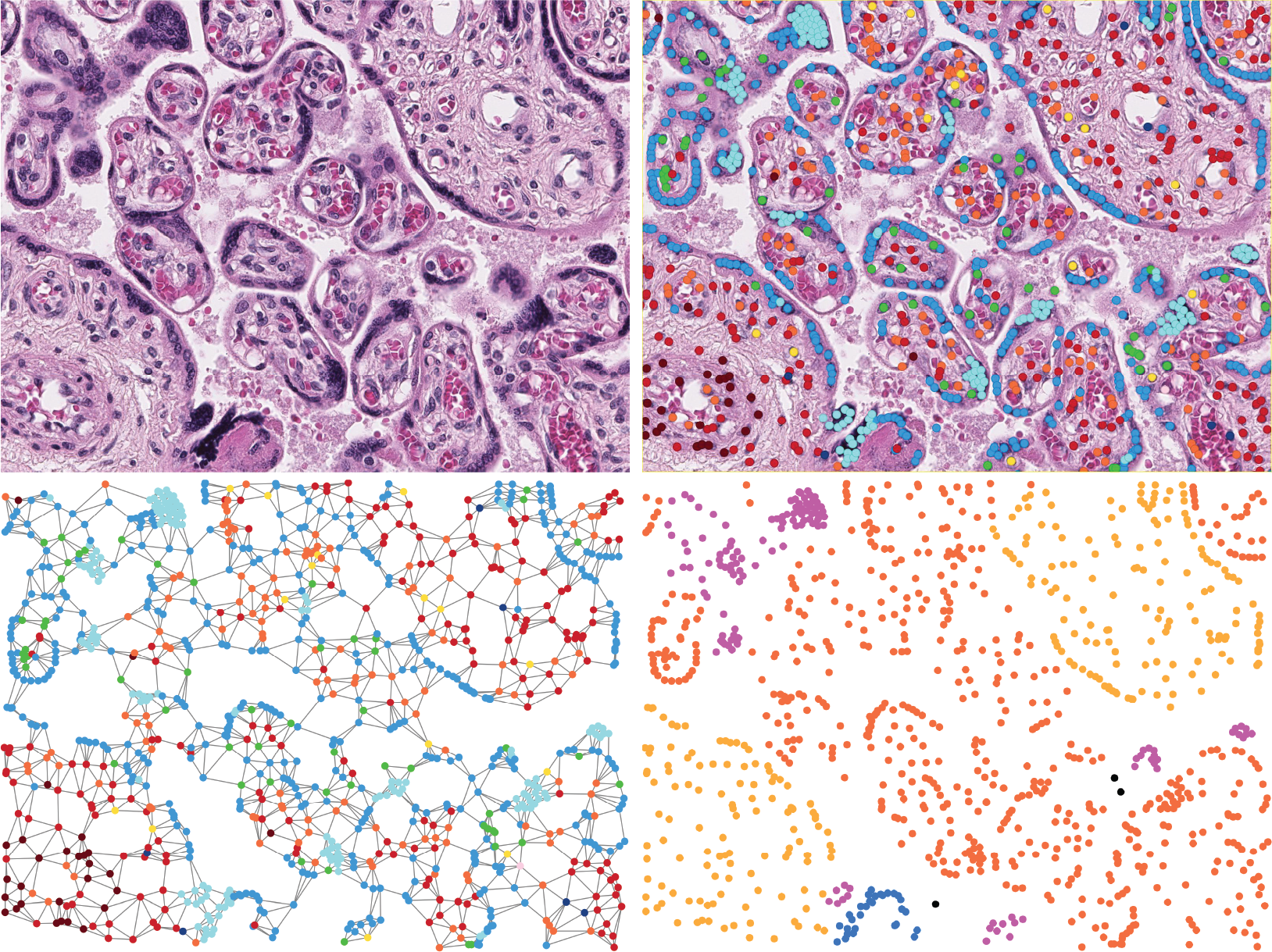}
  \caption{A 1600x1200 pixel region showing (top left) the raw histology image, (top right) overlaid cell nodes, (bottom left) the constructed intersection cell graph, and (bottom right) corresponding ground truth tissue nodes.}
  \label{fig-2}
\end{figure}

The dataset we introduce for this task, \emph{Placenta}, consists of two combined cell graphs constructed across two healthy, term placenta hematoxylin and eosin stained WSIs from two institutes (Table \ref{dataset-characteristics}). Ground truth labels were created by the primary author and validated by four practising perinatal pathologists (details in Vanea et al. \cite{vanea_happy_nodate}). See Figure \ref{fig-1} for a high level visualisation of all cell graph nodes and labelled ground truth nodes and Figure \ref{fig-2} for a representative region. The use of these images was approved and the requirement for consent was waived by local ethics committees at both institutes (approval 0735-18-HMO and 289/T-5). 

We recommend a train, validation, test split (Appendix \ref{splits}) which ensures tissue types have similar distributions across splits. The type of tissue in the placenta can be influenced by macroscopic location (for example, proximity to the chorionic or basal plate), so these dataset split regions are chosen to cover a range of locations and to avoid neighbourhood information leakage.

\paragraph{Nodes and Node Features} We use a prior two-stage deep learning pipeline to identify the centre point of all nuclei across the WSI and classify those nuclei as belonging to one of 11 cell types. The nucleus coordinates define the node coordinates in the graph and the 64-dimension embeddings of the penultimate layer of the cell classifier model are used for the node features. Full details of this pipeline can be found in Vanea et al. \cite{vanea_happy_nodate}.

\paragraph{Edges and Edge Features} We explore three edge-building algorithms which act as a proxy for cellular interaction: KNN \cite{eppstein_nearest-neighbor_1997}, Delaunay Triangulation \cite{guibas_randomized_1992}, and the intersection of the two (Appendix \ref{edges}). Unless stated otherwise, we use the intersection graph with k=5 for all experiments.

Edge construction in the context of cell graphs are representations of the approximate biological likelihood that nearby cells interact as part of the same structure. The Euclidean distance-based edge construction we have explored is one method for this, but alternative approaches for constructing the network are worth considering. We leave this as an open question for the community.

\section{Experiments}
\label{experiments}

\paragraph{Experimental Setup} We first construct a cell graph across all cells in both WSIs. Nodes are masked as either unlabelled, training, validation, or test nodes and only nodes in the training set are used for back propagation. Both cell graphs are combined into one disjointed graph, allowing the sampler to treat it as a dataset of one graph. We restrict our experiments to scalable GNNs as the full graph cannot fit into GPU memory. During training, for \textbf{GraphSAGE} \cite{hamilton_inductive_2017}, \textbf{ClusterGCN} \cite{chiang_cluster-gcn_2019}, \textbf{GraphSAINT} \cite{zeng_graphsaint_2020}, \textbf{ShaDow} \cite{zeng_decoupling_2022} and \textbf{SIGN} \cite{frasca_sign_2020}, node embeddings are sampled or precomputed according to original implementations. For models introduced without a sampling technique (\textbf{GAT} \cite{velickovic_graph_2018} and \textbf{GATv2} \cite{brody_how_2022}), we use ClusterGCN subgraph sampling (ClusterGAT and ClusterGATv2). During inference, all neighbourhood nodes are used for aggregation. We additionally compare against two baselines, an \textbf{MLP on node embeddings} and a \textbf{randomly initialised GraphSAGE}.

\paragraph{Hyperparameters} Models are trained for inductive learning fully-batched to 1000 epochs with early stopping on validation data (evaluated every 100 epochs), weighted cross-entropy loss, Adam optimiser \cite{kingma_adam_2014}, and a 0.001 learning rate. Models have 256 hidden units per layer, a batch normalisation layer following each graph convolutional layer (except for ShaDow), and at most 16 graph convolutional layers each with a 0.5 dropout and relu activations. For GraphSage, ClusterGAT, and ClusterGATv2, 16 layer models could not fit into GPU memory even with reduced sampling and batch sizes and, as such, fewer layers were used. Further hyperparameter details can be found in the supplement (Appendix \ref{hps}).

Note that the goal is not to find the best performing architecture for this dataset, but instead to show the problems existing architectures have on histology cell graphs. As such, we explore hyperparameters to the extent that we believe a fair comparison can be made between models but we do not perform an exhaustive search. This has the added benefit of avoiding overfitting hyperparameters to these two placenta WSIs, which only represent a small facet of the histology domain.

\paragraph{Hardware, Software, and Libraries} Experiments were run on one NVIDIA RTX8000 on an internal cluster. Code was written in Python (v3.7.2) using PyTorch (v1.9.0)\cite{paszke_pytorch_2019} and PyTorch Geometric (v2.0.4)\cite{fey_fast_2019}. WSI visualisation and ground truth annotations were made using QuPath (v0.3.1)\cite{peter_bankhead_qupath_2017}. Code, cell graph dataset, data splits, and experiment instructions may be found at (\url{https://github.com/Nellaker-group/placenta}) under a MIT licence.

\label{code}

\section{Results}
\label{results}

Table \ref{model-results} shows the test mean accuracies and standard deviations of models over 5 random weight initializations. The randomly initialised baseline performs equivalent to random chance (ROC AUC ~0.5) and GNN models outperform the MLP baseline. We found ClusterGAT and ClusterGATv2 were unstable to train with more layers and this might contribute to their lower performance score. All scalable GNN architectures: GraphSAGE, ClusterGCN, GraphSAINT, ShaDow, and SIGN are within 2\% mean accuracy of each other, with no models surpassing 65\% accuracy. These results highlight the need for new GNN approaches for cell graphs in histology.

\begin{table}
  \caption{Mean model performance and standard deviations over 5 random weight initialisation. \textbf{Bold} indicates the best performing model for that metric.}
  \label{model-results}
  \centering
  \begin{tabular}{llll}
    \toprule
    & \textbf{Accuracy} & \textbf{Top 2 Accuracy} & \textbf{ROC AUC} \\
    \midrule
    \textbf{Random Baseline} &  5.56$\pm$6.24 & 14.88$\pm$11.35 & 0.459$\pm$0.057 \\
    \textbf{MLP Baseline} &     47.98$\pm$0.79 & 75.22$\pm$0.92 & 0.750$\pm$0.003 \\
    \textbf{GraphSAGE-mean} &   \textbf{64.88$\pm$0.43} & \textbf{88.94$\pm$0.38} & 0.883$\pm$0.005 \\
    \textbf{ClusterGCN} &       64.24$\pm$1.21 & 88.26$\pm$0.82 & 0.882$\pm$0.006 \\
    \textbf{GraphSAINT-rw} &    63.94$\pm$0.23 & 87.86$\pm$0.15 & \textbf{0.895$\pm$0.002} \\
    \textbf{SIGN} &             64.77$\pm$0.43 & 88.32$\pm$0.42 & 0.886$\pm$0.002 \\
    \textbf{ShaDow} &           63.04$\pm$0.77 & 86.88$\pm$0.74 & 0.863$\pm$0.008 \\
    \textbf{ClusterGAT} &       58.07$\pm$0.61 & 83.43$\pm$0.96 & 0.851$\pm$0.002 \\
    \textbf{ClusterGATv2} &     57.07$\pm$0.65 & 83.21$\pm$0.55 & 0.854$\pm$0.005\\
    \bottomrule
  \end{tabular}
\end{table}

In the supplement, we describe features which the models struggled with and we take a fast, high-performing model (ClusterGCN) to explore some of these in more detail (Appendix \ref{class-imabalance}, \ref{overfitting}, \ref{receptive-field}). We show how class imbalance affects performance, how model size leads to overfitting, and the importance of a large enough receptive field. We additionally compare model performance using all edge-building algorithms (Appendix \ref{edges}), using one hot encoded cell type node features (Appendix \ref{cell-features}), and using random node data splits (Appendix \ref{splits}).

\section{Limitations}
\label{limitations}

Our cell graph dataset represents a small facet of the histology domain; two placenta WSIs from two institutes. WSIs from more placentas and other laboratories will be needed to validate the generalisability of models designed using this data. Likewise, the dataset is placenta-specific, so whilst there is a benefit to using the weights of placenta models for pretraining, the development of GNN tissue classifiers for other organs will require organ-specific cell and tissue ground truth data. Cross- or multi-organ self-supervised approaches could be developed from our dataset but it is not something we have explored. Regarding model comparison, we have purposely limited the scope of our hyperparameter search. It is possible that a different combination of hyperparameters and/or additional sampling, normalisation, and regularisation, would have yielded better performance. We release all data necessary for building cell graphs across our WSIs but not the raw images or pipeline needed to derive the cell features from scratch as this was out of scope for this work. Finally, it should be noted that uncertainty in ground truth annotations may limit model performance on the dataset, however label uncertainty is inherent to the biological problem and unavoidable in this domain.

\section{Conclusion}

We introduce a new inductive node classification benchmark dataset, classifying microanatomical tissue structures from cell graphs across placenta histology WSIs. We show how GNN architectures perform on this task and highlight the qualities unique to this data domain which present a challenge for existing models. We hope that this dataset facilitates the creation of new model architectures to overcome these challenges.

\begin{ack}

Claudia Vanea and Jonathan Campbell are supported by the EPSRC Center for Doctoral Training in Health Data Science (EP/S02428X/1). Cecilia M. Lindgren is supported by the Li Ka Shing Foundation, NIHR Oxford Biomedical Research Centre, Oxford, NIH (1P50HD104224-01), Gates Foundation (INV-024200), and a Wellcome Trust Investigator Award (221782/Z/20/Z). Triin Laisk is funded by the European Regional Development Fund and the programme Mobilitas Pluss (MOBTP155) and the Estonian Research Council grant PSG776. The computational aspects of this research were supported by the Wellcome Trust Core Award Grant Number 203141/Z/16/Z and the NIHR Oxford BRC. The views expressed are those of the author(s) and not necessarily those of the NHS, the NIHR or the Department of Health.

\end{ack}

\pagebreak
\medskip

{
\small
\bibliography{references.bib}
\bibliographystyle{abbrvnat}

}

\pagebreak
\section*{Checklist}

\begin{enumerate}

\item For all authors...
\begin{enumerate}
  \item Do the main claims made in the abstract and introduction accurately reflect the paper's contributions and scope?
    \answerYes{We claim to introduce a new dataset benchmark in histology cell graphs and show how existing GNN models perform on this data.}
  \item Did you describe the limitations of your work?
    \answerYes{See Section~\ref{limitations}.}
  \item Did you discuss any potential negative societal impacts of your work?
    \answerYes{See Appendix~\ref{negative-societal-impacts}.}
  \item Have you read the ethics review guidelines and ensured that your paper conforms to them?
    \answerYes{}
\end{enumerate}

\item If you are including theoretical results...
\begin{enumerate}
  \item Did you state the full set of assumptions of all theoretical results?
    \answerNA{}
        \item Did you include complete proofs of all theoretical results?
    \answerNA{}
\end{enumerate}

\item If you ran experiments...
\begin{enumerate}
  \item Did you include the code, data, and instructions needed to reproduce the main experimental results (either in the supplemental material or as a URL)?
    \answerYes{See Section~\ref{code} and at (\url{https://github.com/Nellaker-group/placenta}).} 
  \item Did you specify all the training details (e.g., data splits, hyperparameters, how they were chosen)?
    \answerYes{See Section~\ref{dataset-description}, ~\ref{experiments} and Appendix~\ref{splits}, ~\ref{hps}.}
        \item Did you report error bars (e.g., with respect to the random seed after running experiments multiple times)?
    \answerYes{See Section~\ref{results}. Experiments were run with 5 random seeds.}
        \item Did you include the total amount of compute and the type of resources used (e.g., type of GPUs, internal cluster, or cloud provider)?
    \answerYes{See Section~\ref{code} and Appendix ~\ref{CO2}.}
\end{enumerate}

\item If you are using existing assets (e.g., code, data, models) or curating/releasing new assets...
\begin{enumerate}
  \item If your work uses existing assets, did you cite the creators?
    \answerYes{We existing GNN models and code from PyTorch Geometric and we cite the creators}
  \item Did you mention the license of the assets?
    \answerYes{See Section~\ref{code}.}
  \item Did you include any new assets either in the supplemental material or as a URL?
    \answerYes{See Section~\ref{dataset-description} and at (\url{https://github.com/Nellaker-group/placenta}).}
  \item Did you discuss whether and how consent was obtained from people whose data you're using/curating?
    \answerYes{See Section~\ref{dataset-description}. Data was obtained without consent with approval from local ethics committees for research use.}
  \item Did you discuss whether the data you are using/curating contains personally identifiable information or offensive content?
    \answerYes{See Section~\ref{dataset-description} and ~\ref{limitations}.}
\end{enumerate}

\item If you used crowdsourcing or conducted research with human subjects...
\begin{enumerate}
  \item Did you include the full text of instructions given to participants and screenshots, if applicable?
    \answerNA{}
  \item Did you describe any potential participant risks, with links to Institutional Review Board (IRB) approvals, if applicable?
    \answerNA{}
  \item Did you include the estimated hourly wage paid to participants and the total amount spent on participant compensation?
    \answerNA{}
\end{enumerate}

\end{enumerate}


\pagebreak
\appendix

\section{Placenta Cell and Tissue Types}
\label{cell-tissue-types}

11 placenta cells used in node features:
\begin{itemize}
\item SYN: Syncytiotrophoblast
\item CYT: Cytotrophoblast
\item KNT: Syncytial Knot
\item EVT: Extravillus Trophoblast
\item FIB: Fibroblast
\item HOF: Hofbauer Cell
\item VEN: Vascular Endothelial
\item VMY: Vascular Myocyte
\item MES: Mesenchymal Cell
\item MAT: Maternal Decidua
\item WBC: Leukocyte
\end{itemize}

9 placenta tissue types used as node labels:
\begin{itemize}
\item Sprout: Villus Sprout
\item TVilli: Terminal Villi
\item MIVilli: Mature Intermediate Villi
\item SVilli: Stem Villi
\item AVilli: Anchoring Villi
\item Chorion: Chorionic Plate
\item Maternal: Basal Plate and Septum
\item Fibrin: Fibrin
\item Avacular: Avascular Villi
\end{itemize}

\pagebreak
\section{Cell Types in Node Features and Ground Truth Tissue Distributions}
\label{distributions}

\begin{figure}[h]
  \centering
  \includegraphics[scale=0.90]{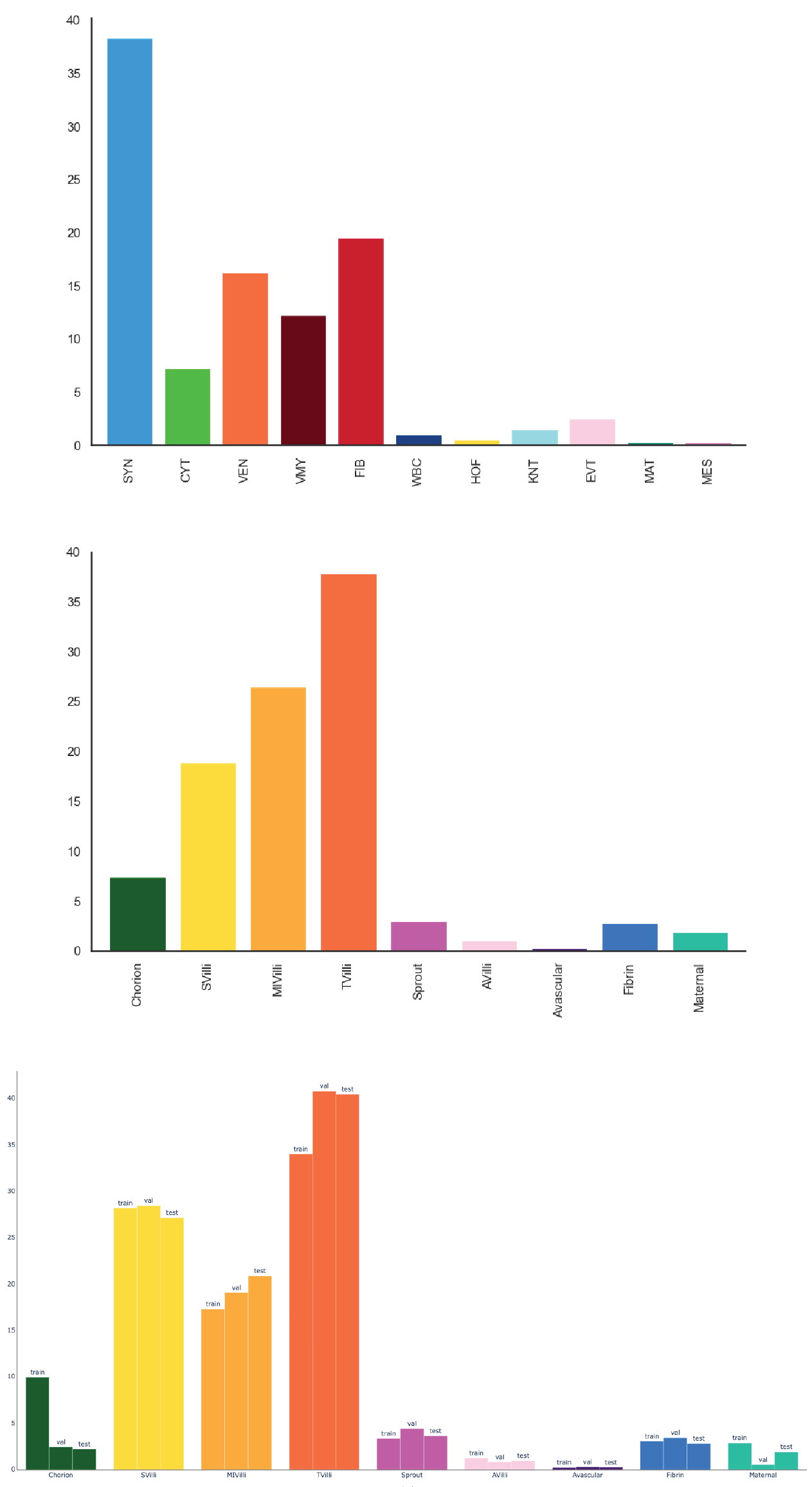}
  \caption{Cell type proportions in node features (top), tissue label proportions in all labelled data (middle), and distribution of tissue classes within dataset splits (bottom).}
\end{figure}

\pagebreak
\section{Data Splits and the Effect of Random Node Splits}
\label{splits}

\begin{figure}[h]
  \centering
  \includegraphics[scale=0.70]{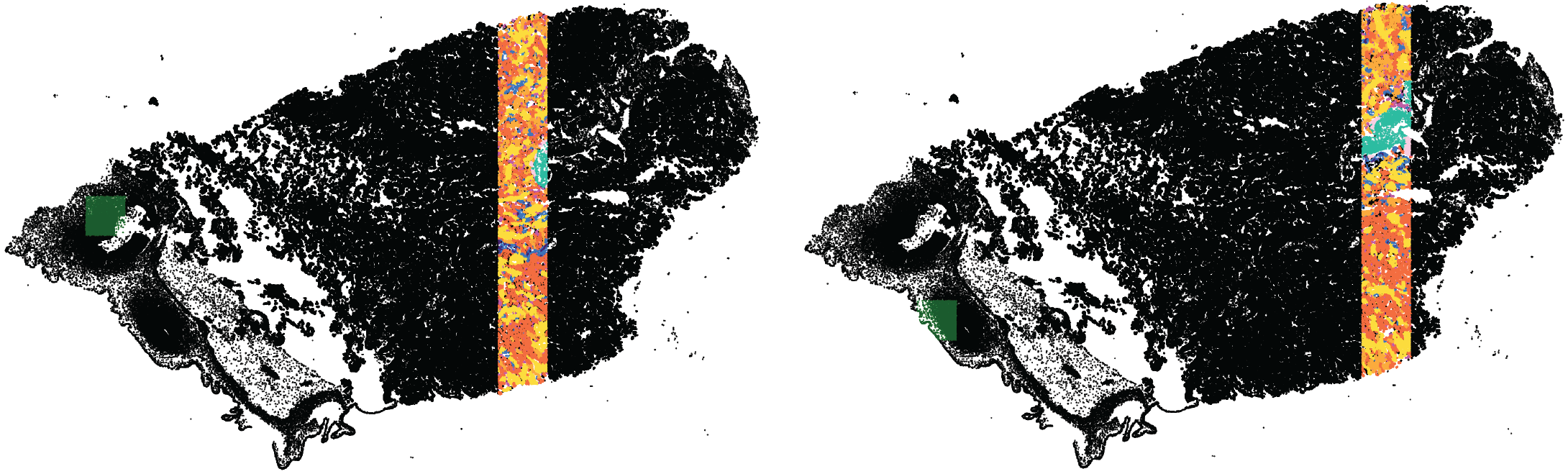}
  \caption{Validation region (left) and test region (right).}
\end{figure}

\begin{table}[h]
  \caption{Distribution of class labels across dataset splits.}
  \label{data-splits-table}
  \centering
  \begin{tabular}{llll}
    \toprule
    \textbf{Class Label} & \textbf{Training} & \textbf{Validation} & \textbf{Test}\\
    \midrule
    \textbf{Sprout} & 15,545 (50\%) & 9,066 (29\%) & 6,441 (21\%)\\
    \textbf{TVilli} & 159,464 (50\%) & 84,052 (27\%) & 72,402 (23\%)\\
    \textbf{MIVilli} & 80,954 (51\%) & 39,296 (25\%) & 37,332 (24\%)\\
    \textbf{SVilli} & 132,138 (55\%) & 58,658 (25\%) & 48,568 (20\%)\\
    \textbf{AVilli} & 5,769 (64\%) & 1,634 (18\%) & 1,654 (18\%)\\
    \textbf{Chorion} & 46,584 (84\%) & 4,942 (9\%) & 3,937 (7\%)\\
    \textbf{Maternal} & 13,329 (75\%) & 1,029 (6\%) & 3,359 (19\%)\\
    \textbf{Fibrin} & 14,195 (54\%) & 7,047 (27\%) & 5,010 (19\%)\\
    \textbf{Avacular} & 891 (49\%) & 526 (29\%) & 392 (22\%)\\
    \bottomrule
  \end{tabular}
\end{table}

\begin{figure}[h]
  \centering
  \includegraphics[scale=0.60]{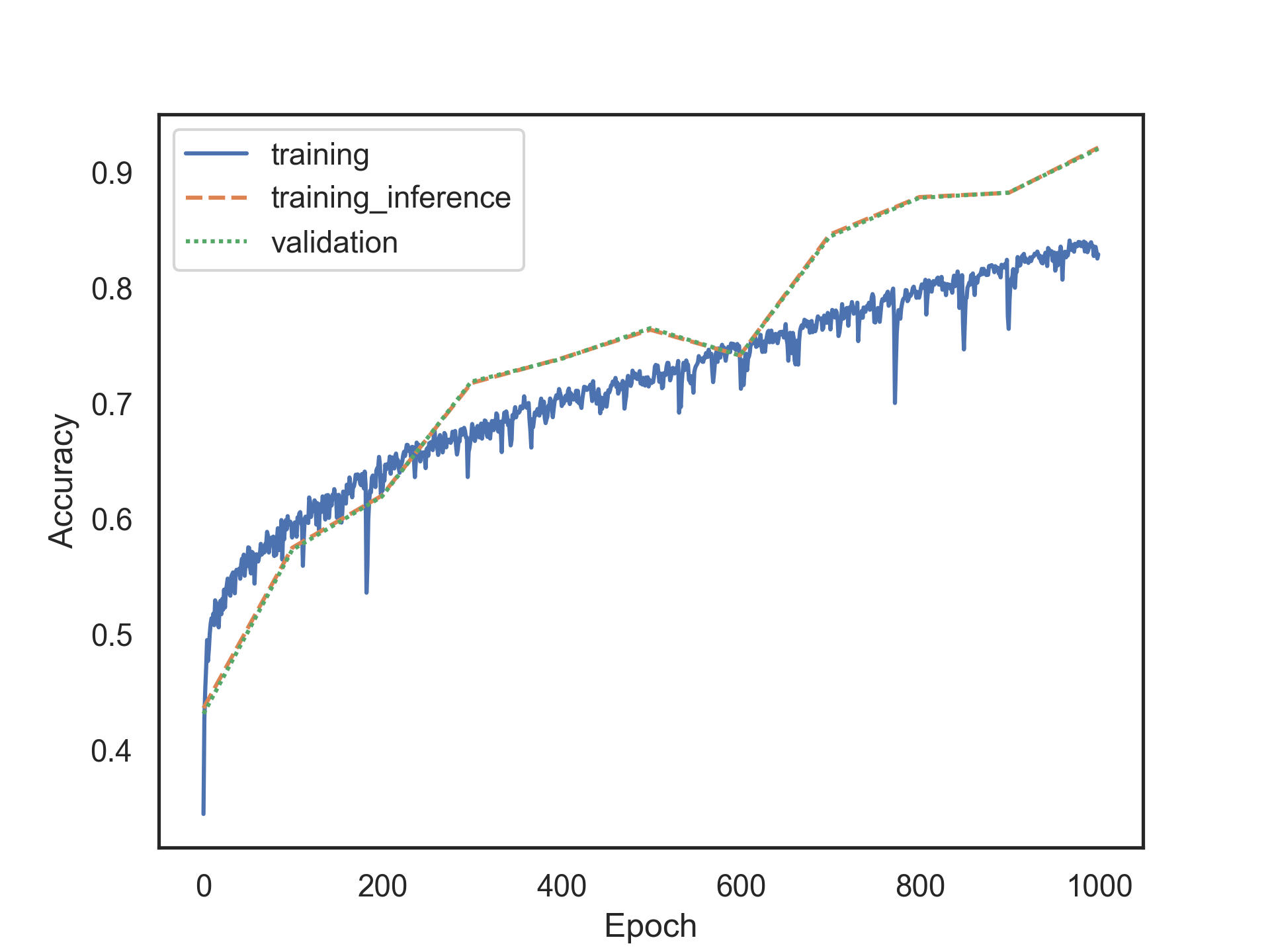}
  \caption{Accuracy curves during training using random node splits of 70\%/15\%/15\%. Validation performance is identical to training performance using the same inference sampling method, showing information leakage across splits.}
\end{figure}

\pagebreak
\section{Edge Construction and Graph Differences}
\label{edges}

\begin{table}[h]
  \caption{Graph Characteristics of KNN, Delaunay, and intersection graphs.}
  \label{all-graph-characteristics}
  \centering
  \begin{tabular}{llllll}
    \toprule
    & \textbf{Nodes} & \textbf{Edges} & \textbf{Labels} & \textbf{Homophily} & \textbf{Degree} \\
    \midrule
    \textbf{Intersection Graph 1} & 1,159,186 & 2,603,797 & 703,165 & 0.9868 & 5.4925 \\
    \textbf{Intersection Graph 2} & 1,236,561 & 2,882,292 & 96,580  & 0.9984 & 5.6618 \\
    \textbf{KNN Graph 1} & 1,159,186 & 8,398,932 & 703,165 & 0.9854 & 8.2455 \\
    \textbf{KNN Graph 2} & 1,236,561 & 8,870,066 & 96,580 & 0.7286 & 8.1732 \\
    \textbf{Delaunay Graph 1} & 1,159,186 & 3,477,526 & 703,165 & 0.9522 & 6.9999 \\
    \textbf{Delaunay Graph 2} & 1,236,561 & 3,709,652 & 96,580 & 0.7217 & 6.9999 \\
    \bottomrule
  \end{tabular}
\end{table}

\begin{figure}[h]
  \centering
  \includegraphics[scale=0.70]{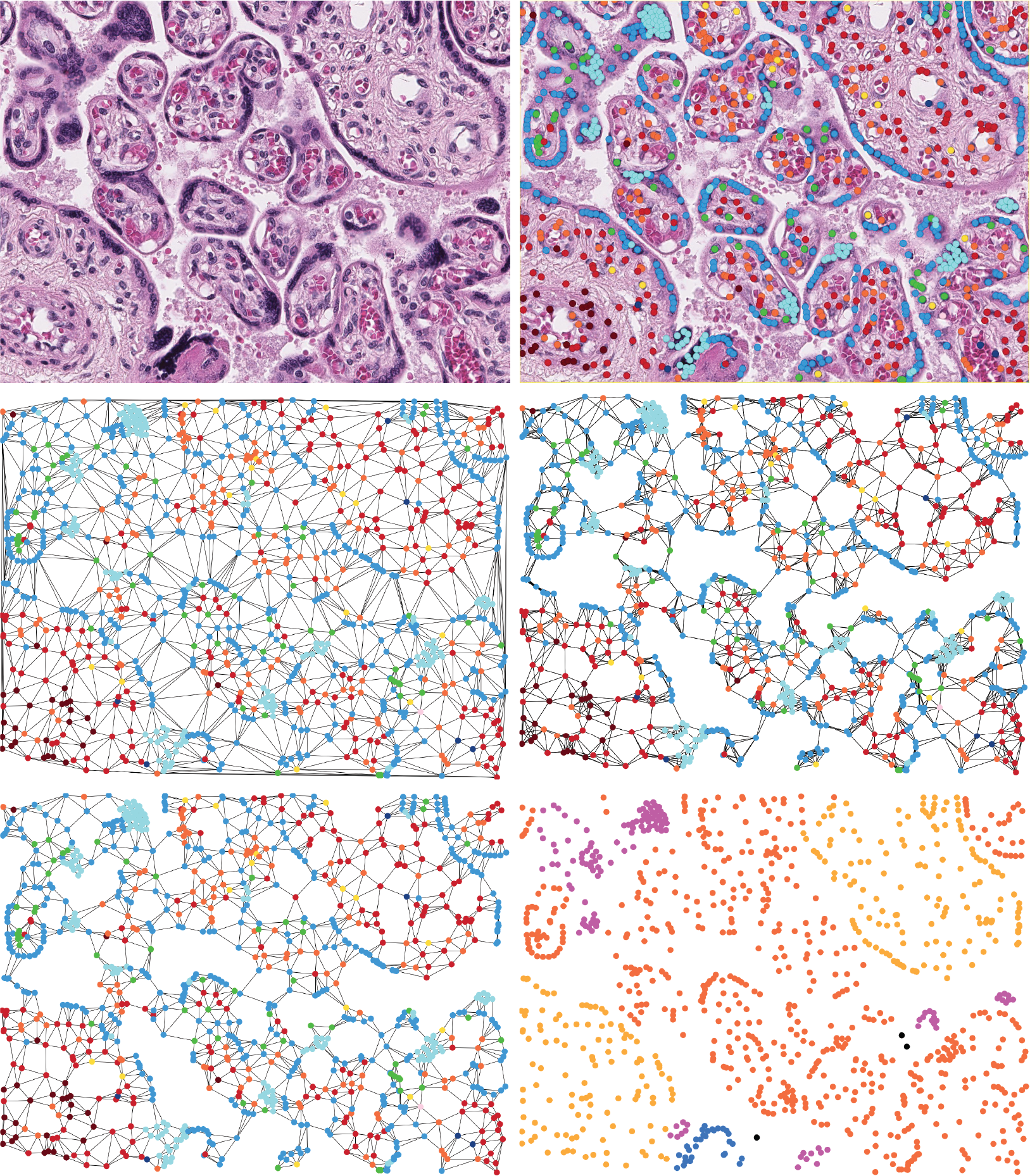}
  \caption{From top left to bottom right, the raw histology, histology with cell overlay, Delaunay Triangulation edges, KNN edges, intersection edges, and ground truth tissue labels.}
\end{figure}

\begin{table}[h]
  \caption{ClusterGCN performance with a KNN graph where k=5 and a Delaunay graph.}
  \label{model-results-edges}
  \centering
  \begin{tabular}{llll}
    \toprule
    & \textbf{Accuracy} & \textbf{Top 2 Accuracy} & \textbf{ROC AUC} \\
    \midrule
    \textbf{KNN} & 0.654 & 0.894 & 0.903 \\
    \textbf{Delaunay} & 0.676 & 0.898 & 0.916 \\
    \bottomrule
  \end{tabular}
\end{table}

\pagebreak
\section{Hyperparameter Details for Each Model Architecture}
\label{hps}

Weighted cross entropy weights: [1, 0.85, 0.9, 10.5, 0.8, 1.3, 5.6, 3, 77]

Additional hyperparameter details and any changes from the default hyperparameters reported in Section \ref{experiments} are reported below.

\paragraph{MLP} 16 Linear layers, 51,200 batch size.
\paragraph{GraphSAGE-mean} 12 SAGEConv layers due to GPU out of memory error with larger models, 32,000 batch size, 10 neighbours per hop, mean aggregation.
\paragraph{ClusterGCN} 16 SAGEConv layers, 200 batch size, and 400 max subgraph size.
\paragraph{GraphSAINT-rw} 16 GraphConv layers, in-node normalisation, 32,000 batch size, 30 steps, 500 sample coverage, random walk sampler.
\paragraph{SIGN} 16 Linear layers, 51,200 batch size, SIGN transform with 16 hops.
\paragraph{ShaDow} 8 SAGEConv layers due to performance drop with larger models, no batch norm layers for the same reason, 4000 batch size, 5 neighbours per hop, hop depth of 6. This smaller receptive field is due to a RAM memory error with a higher depth or more neighbours.
\paragraph{ClusterGAT} 2 GATConv layers due to GPU out of memory error and performance drop with larger models, 400 batch size, 400 max subgraph size, 4 attention heads with 0.25 dropout.
\paragraph{ClusterGATv2} 2 GATv2Conv layers due to GPU out of memory error and performance drop with larger models, 200 batch size, 400 max subgraph size, 4 attention heads with 0.25 dropout.

\section{The Effect of Class imbalance on Performance}
\label{class-imabalance}

\begin{figure}[h]
  \centering
  \includegraphics[scale=0.70]{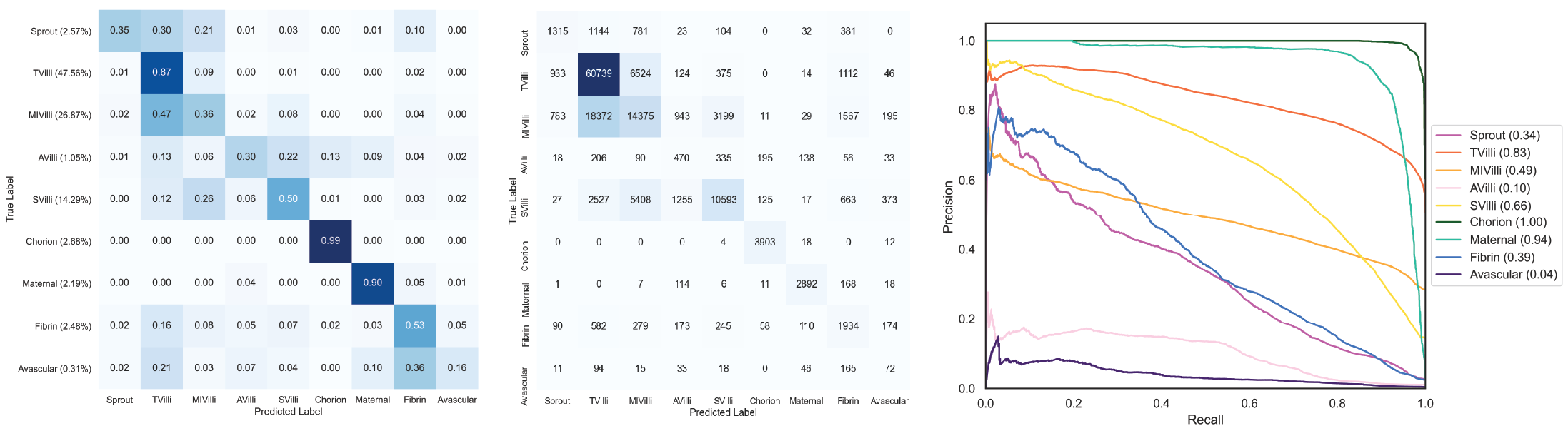}
  \caption{Confusion matrix with values proportional to class imbalance (left), confusion matrix per number of nodes (middle), and precision-recall curves with corresponding AUC-PR values (right). The dataset is highly imbalanced and models perform worse on many minority classes.}
\end{figure}

\pagebreak
\section{The Effect of Model Size on Overfitting to Training Data}
\label{overfitting}

\begin{figure}[h]
  \centering
  \includegraphics[scale=0.71]{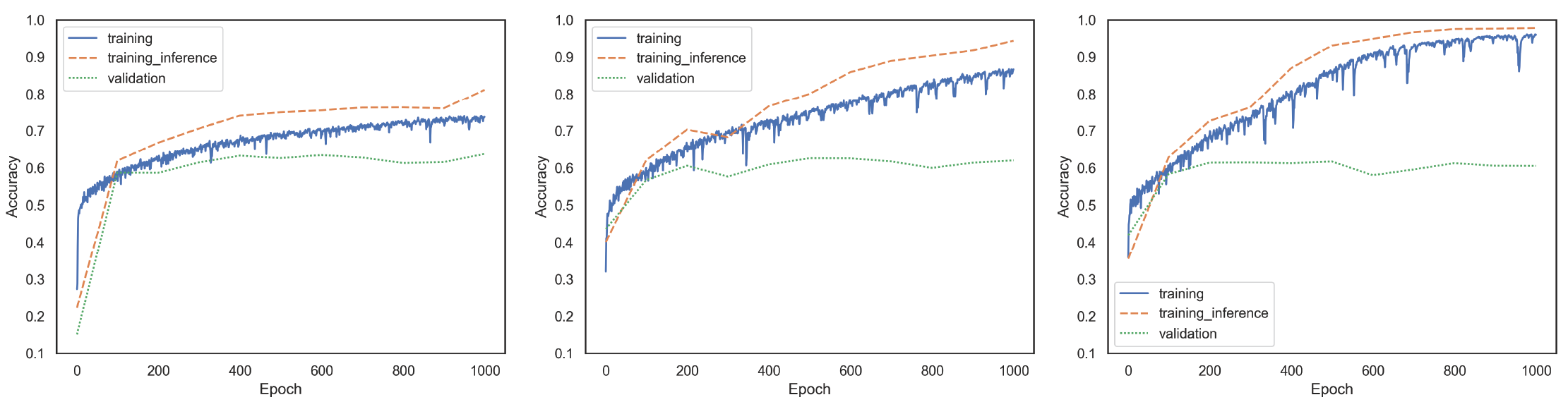}
  \caption{ClusterGCN with 128 hidden units (left), 256 hidden units (middle), and 512 hidden units (right). Too few hidden units and the model does not learn the training data, too many and the model overfits.}
\end{figure}

\section{The Effect of Receptive Field Sizes}
\label{receptive-field}

\begin{table}[h]
  \caption{ClusterGCN performance with varied receptive fields. Receptive fields are varied by changing cluster subgraph sizes and the number of model layers.}
  \label{model-results-field}
  \centering
  \begin{tabular}{llll}
    \toprule
    & \textbf{Accuracy} & \textbf{Top 2 Accuracy} & \textbf{ROC AUC} \\
    \midrule
    \textbf{800 Subgraph 32 Layers} & 0.647 & 0.874 & 0.893 \\
    \textbf{400 Subgraph 16 Layers} & 0.642 & 0.883 & 0.882 \\
    \textbf{200 Subgraph 8 Layers} & 0.623 & 0.871 & 0.876 \\
    \textbf{100 Subgraph 4 Layers} & 0.594 & 0.849 & 0.856 \\
    \bottomrule
  \end{tabular}
\end{table}

\begin{figure}[h]
  \centering
  \includegraphics[scale=1.10]{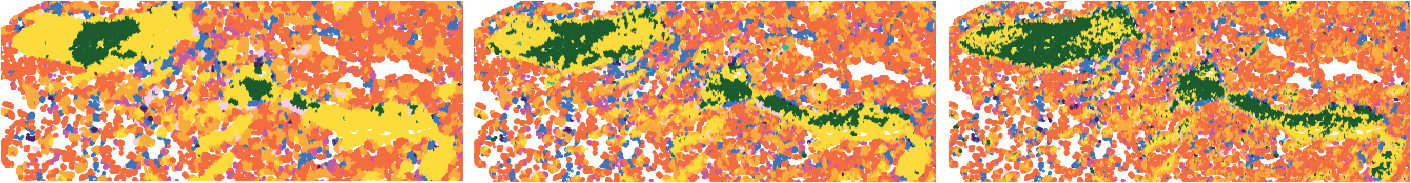}
  \caption{Reducing receptive field sizes (left to right) leads to confused predictions inside large structures. The yellow tissue structure (SVilli) in these images is incorrectly labelled as the green tissue structure (Chorion).}
\end{figure}

\section{The Effect of Cellular Node Features}
\label{cell-features}

\begin{table}[!ht]
  \caption{ClusterGCN performance when trained and evaluated with one hot encoded cell types. Performance is marginally worse than training with 64-dim feature vectors.}
  \label{model-results-celltype-features}
  \centering
  \begin{tabular}{llll}
    \toprule
    & \textbf{Accuracy} & \textbf{Top 2 Accuracy} & \textbf{ROC AUC} \\
    \midrule
    \textbf{One Hot Cell Types} & 0.628 & 0.871 & 0.833 \\
    \bottomrule
  \end{tabular}
\end{table}

\pagebreak
\section{Visualisation of Model Predictions}
\label{vis-umap}

\begin{figure}[h]
  \centering
  \includegraphics[scale=0.92]{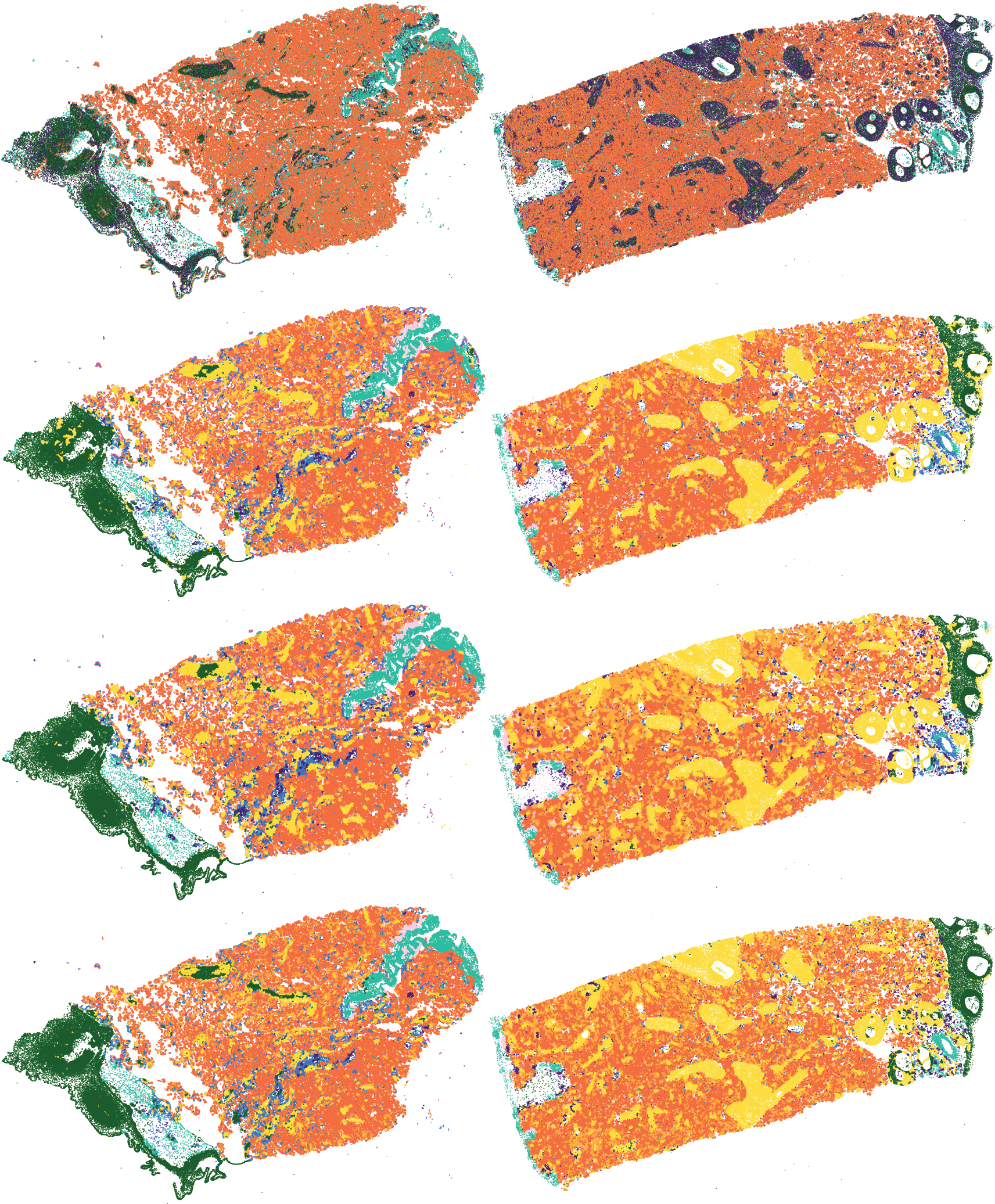}
  \caption{From top to bottom, MLP, GraphSAGE-mean, ClusterGCN, and GraphSAINT-rw tissue predictions across both WSIs.}
\end{figure}

\pagebreak

\begin{figure}[h]
  \centering
  \includegraphics[scale=0.92]{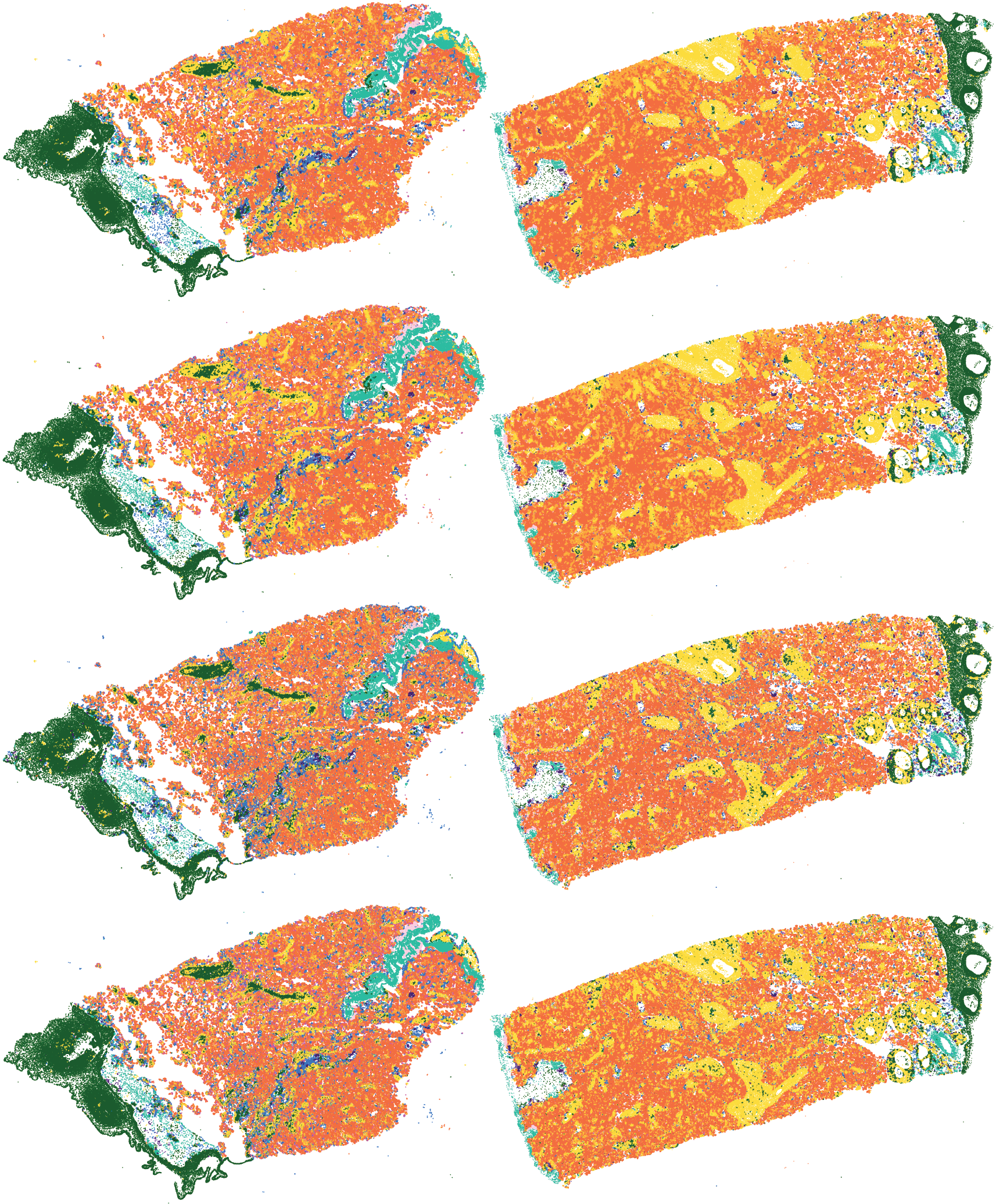}
  \caption{From top to bottom, SIGN, ShaDow, ClusterGAT, and ClusterGATv2 tissue predictions across both WSIs.}
\end{figure}

\pagebreak
\section{Potential Negative Societal Impacts}
\label{negative-societal-impacts}

The development of similar models in the context of histological analyses could, if in future fully realised and implemented across healthcare service settings, negatively impact the work of pathologists who do not have access to the models: “AI won’t replace radiologists, but radiologists who use AI will replace radiologists who don’t,” (Curtis Langlotz, 2019, \href{https://www.nature.com/articles/d41586-019-03847-z}{Nature, Innovations In}). As with the uptake of any new technology, there would be a ‘growing pains’ period which could increase the time between tissue sample collection and diagnosis. Slides will have to be digitised and model inference will have to be integrated, causing a shift in existing clinical and histological preparation workflows. We must also consider that any models developed for histology analysis could cause serious patient harm if, in the extreme case, they are incorrectly validated and quality controlled, or they have an undetected bias which influences diagnoses. If we assume models perform as well or better than clinicians, but their outputs are not clearly expressed in clinical language and/or there is a lack of trust in these automated, uninterpretable methods, then there may be resulting uncertainty in diagnosis and consequently higher referral rates. Conversely, if automated histology analysis becomes popular, for organs unlike the placenta which require biopsies to be sampled, the rate of biopsies could harmfully increase. Finally, the publication of this dataset could facilitate the development of new GNN architectures that themselves could confer negative impacts in other domains.

\section{CO2 Emission Related to Experiments}
\label{CO2}

Experiments were conducted using a private infrastructure, which has a carbon efficiency of 0.432 kgCO$_2$eq/kWh. An approximate cumulative of 200 hours of computation was performed on hardware of type RTX 8000 (TDP of 260W).

Total emissions are estimated to be 22.46 kgCO$_2$eq of which 0 percents were directly offset.

Estimations were conducted using the \href{https://mlco2.github.io/impact#compute}{MachineLearning Impact calculator} presented in \cite{lacoste_quantifying_2019}.

\end{document}